\documentclass{article}

\usepackage[final]{nips_2016}
\usepackage{algorithm}
\usepackage{algorithmic}
\usepackage{array}
\usepackage{pgfplots}
\usepackage{natbib}

\usepackage[utf8]{inputenc} 
\usepackage[T1]{fontenc}    
\usepackage{hyperref}       
\usepackage{url}            
\usepackage{booktabs}       
\usepackage{amsfonts}       
\usepackage{nicefrac}       
\usepackage{microtype}      
\usepackage[inline]{enumitem}

\newcommand{\ie}{\emph{i.e.}}
\newcommand{\eg}{\emph{e.g.}}

\title{MS MARCO: A Human Generated MAchine Reading COmprehension Dataset}

\author{
\textbf{Payal Bajaj, Daniel Campos, Nick Craswell, Li Deng, Jianfeng Gao,} \\
\textbf{Xiaodong Liu, Rangan Majumder, Andrew McNamara, Bhaskar Mitra, Tri Nguyen,} \\
\textbf{Mir Rosenberg, Xia Song, Alina Stoica, Saurabh Tiwary, and Tong Wang} \\
     {Microsoft AI \& Research}
}

\begin{document}
\maketitle

\begin{abstract}
We introduce a large scale MAchine Reading COmprehension dataset, which we name MS MARCO.
The dataset comprises of 1,010,916 anonymized questions---sampled from Bing's search query logs---each with a human generated answer and 182,669 completely human rewritten generated answers.
In addition, the dataset contains 8,841,823 passages---extracted from 3,563,535 web documents retrieved by Bing---that provide the information necessary for curating the natural language answers.
A question in the MS MARCO dataset may have multiple answers or no answers at all.
Using this dataset, we propose three different tasks with varying levels of difficulty:
\begin{enumerate*}[label=(\roman*)]
    \item predict if a question is answerable given a set of context passages, and extract and synthesize the answer as a human would
    \item generate a well-formed answer (if possible) based on the context passages that can be understood with the question and passage context, and finally
    \item rank a set of retrieved passages given a question.
\end{enumerate*}
The size of the dataset and the fact that the questions are derived from real user search queries distinguishes MS MARCO from other well-known publicly available datasets for machine reading comprehension and question-answering.
We believe that the scale and the real-world nature of this dataset makes it attractive for benchmarking machine reading comprehension and question-answering models.
\end{abstract}

\section{Introduction}
\label{sec:intro}

Building intelligent agents with machine reading comprehension (MRC) or open-domain question answering (QA) capabilities using real world data is an important goal of artificial intelligence.
Progress in developing these capabilities can be of significant consumer value if employed in automated assistants---\eg, Cortana \citep{web:cortana}, Siri \citep{web:siri}, Alexa \citep{web:alexa}, or Google Assistant \citep{web:google-assistant}---on mobile devices and smart speakers, such as Amazon Echo \citep{web:echo}.
Many of these devices rely heavily on recent advances in speech recognition technology powered by neural models with deep architectures \citep{Hinton2012,Dahl2012}.
The rising popularity of spoken interfaces makes it more attractive for users to use natural language dialog for question-answering and information retrieval from the web as opposed to viewing traditional search result pages on a web browser \citep{gaosurvey}.
Chatbots and other messenger based intelligent agents are also becoming popular in automating business processes---\eg, answering customer service requests.
All of these scenarios can benefit from fundamental improvements in MRC models.
However, MRC in the wild is extremely challenging.
Successful MRC systems should be able to learn good representations from raw text, infer and reason over learned representations, and finally generate a summarized response that is correct in both form and content.

The public availability of large datasets has been instrumental in many AI research breakthroughs \citep{wissner2016datasets}.
For example, ImageNet's \citep{deng:ImageNet} release of 1.5 million labeled examples with 1000 object categories led to the development of object classification models that perform better than humans on the ImageNet task \citep{he:ResNet}.
Similarly, the large speech database collected over 20 years by DARPA enabled new breakthroughs in speech recognition performance from deep learning models \cite {DengHuang2004}.
Several MRC and QA datasets have also recently emerged.
However, many of these existing datasets are not sufficiently large to train deep neural models with large number of parameters.
Large scale existing MRC datasets, when available, are often synthetic.
Furthermore, a common characteristic, shared by many of these datasets, is that the questions are usually generated by crowd workers based on provided text spans or documents.
In MS MARCO, in contrast, the questions correspond to actual search queries that users submitted to Bing, and therefore may be more representative of a ``natural'' distribution of information need that users may want to satisfy using, say, an intelligent assistant.

Real-world text is messy: they may include typos or abbreviations---and transcription errors in case of spoken interfaces.
The text from different documents may also often contain conflicting information.
Most existing datasets, in contrast, often contain high-quality stories or text spans from sources such as Wikipedia.
Real-world MRC systems should be benchmarked on realistic datasets where they need to be robust to noisy and problematic inputs.

Finally, another potential limitation of existing MRC tasks is that they often require the model to operate on a single entity or a text span.
Under many real-world application settings, the information necessary to answer a question may be spread across different parts of the same document, or even across multiple documents.
It is, therefore, important to test an MRC model on its ability to extract information and support for the final answer from multiple passages and documents.

In this paper, we introduce Microsoft MAchine Reading Comprehension (MS MARCO)---a large scale real-world reading comprehension dataset---with the goal of addressing many of the above mentioned shortcomings of existing MRC and QA datasets.
The dataset comprises of anonymized search queries issued through Bing or Cortana.
We annotate each question with segment information as we describe in Section \ref{sec:data}.
Corresponding to each question, we provide a set of extracted passages from documents retrieved by Bing in response to the question.
The passages and the documents may or may not actually contain the necessary information to answer the question.
For each question, we ask crowd-sourced editors to generate answers based on the information contained in the retrieved passages.
In addition to generating the answer, the editors are also instructed to mark the passages containing the supporting information---although we do not enforce these annotations to be exhaustive.
The editors are allowed to mark a question as unanswerable based on the passages provided.
We include these unanswerable questions in our dataset because we believe that the ability to recognize insufficient (or conflicting) information that makes a question unanswerable is important to develop for an MRC model.
The editors are strongly encouraged to form answers in complete sentences.
In total, the MS MARCO dataset contains 1,010,916 questions, 8,841,823 companion passages extracted from 3,563,535 web documents, and 182,669 editorially generated answers.
Using this dataset, we propose three different tasks with varying levels of difficulty:
\begin{enumerate}[label=(\roman*)]
    \item Predict if a question is answerable given a set of context passages, and extract relevant information and synthesize the answer.
    \item Generate a well-formed answer (if possible) based on the context passages that can be understood with the question and passage context.
    \item Rank a set of retrieved passages given a question.
\end{enumerate}

We describe the dataset and the proposed tasks in more details in the rest of this paper and present some preliminary benchmarking results on these tasks.

\section{Related work}
\label{sec:related}

Machine reading comprehension and open domain question-answering are challenging tasks \citep{weston:babi}.
To encourage more rapid progress, the community has made several different datasets and tasks publicly available for benchmarking.
We summarize some of them in this section.

\begin{table}
\caption{Comparison of MS MARCO and some of the other MRC datasets.}
\begin{center}
\resizebox{\textwidth}{!}{
\setlength{\tabcolsep}{4pt}
\begin{tabular}{l l l l l l} \\
\textbf{Dataset} & \textbf{Segment} & \textbf{Question Source} & \textbf{Answer} & \textbf{\# Questions} & \textbf{\# Documents} \\
\hline
NewsQA & No & Crowd-sourced & Span of words & 100k & 10k \\
DuReader & No & Crowd-sourced & Human generated & 200k & 1M \\
NarrativeQA & No & Crowd-sourced & Human generated & 46,765 & 1,572 stories  \\
SearchQA & No & Generated & Span of words & 140k & 6.9M passages \\
RACE & No & Crowd-sourced & Multiple choice & 97k & 28k \\
ARC & No & Generated & Multiple choice & 7,787 & 14M sentences\\
SQuAD & No & Crowd-sourced & Span of words & 100K & 536\\
MS MARCO & Yes & User logs & Human generated & 1M & 8.8M passages, 3.2m docs.\\
\hline
\end{tabular}}
\label{table:ds-comparison}
\end{center}
\end{table}

\begin{description}[nosep,leftmargin=0.13in]

    \item [The Stanford Question Answering Dataset (SQuAD)] \cite {raj:squad}  consists of 107,785 question-answer pairs from 536 articles, where each answer is a text span.
    The key distinction between SQUAD and MS MARCO are:
    \begin{enumerate}
        \item The MS MARCO dataset is more than ten times larger than SQuAD---which is an important consideration if we want to benchmark large deep learning models \citep{web:MinimumExamplesDeepLearning}.
        \item The questions in SQuAD are editorially generated based on selected answer spans, while in MS MARCO they are sampled from Bing's query logs.
        \item The answers in SQuAD consists of spans of texts from the provided passages while the answers in MS MARCO are editorially generated.
        \item Originally SQuAD contained only answerable questions, although this changed in the more recent edition of the task \citep{rajpurkar2018know}.
    \end{enumerate}
    
    \item[NewsQA] \citep{Trischler2017NewsQAAM} is a MRC dataset with over 100,000 question and span-answer pairs based off roughly 10,000 CNN news articles.
    The goal of the NewsQA task is to test MRC models on reasoning skills---beyond word matching and paraphrasing.
    Crowd-sourced editors created the questions from the title of the articles and the summary points (provided by CNN) without access to the article itself.
    A 4-stage collection methodology was employed to generate a more challenging MRC task.
    More than 44\% of the NewsQA questions require inference and synthesis, compared to SQuAD's 20\%.
    
    \item[DuReader] \citep{He2017DuReaderAC} is a Chinese MRC dataset built with real application data from Baidu search and Baidu Zhidao---a community question answering website.
    It contains 200,000 questions and 420,000 answers from 1,000,000 documents.
    In addition, DuReader provides additional annotations of the answers---labelling them as either fact based or opinionative.
    Within each category, they are further divided into entity, yes/no, and descriptive answers.
    
    \item[NarrativeQA] \citep{Kocisk2017TheNR} dataset contains questions created by editors based on summaries of movie scripts and books.
    The dataset contains about 45,000 question-answer pairs over 1,567 stories, evenly split between books and movie scripts.
    Compared to the news corpus used in NewsQA, the collection of movie scripts and books are more complex and diverse---allowing the editors to create questions that may require more complex reasoning.
    The movie scripts and books are also longer documents than the news or wikipedia article, as is the case with NewsQA and SQuAD, respectively.
    
    \item[SearchQA] \citep{Dunn2017SearchQAAN} takes questions from the American TV quiz show, Jeopardy\footnote{\url{https://www.jeopardy.com/}} and submits them as queries to Google to extract snippets from top 40 retrieved documents that may contain the answers to the questions.
    Document snippets not containing answers are filtered out, leaving more than 140K questions-answer pairs and 6.9M snippets.
    The answers are short exact spans of text averaging between 1-2 tokens.
    MS MARCO, in contrast, focuses more on longer natural language answer generation, and the questions correspond to Bing search queries instead of trivia questions.
    
    \item[RACE] \citep{Lai2017RACELR} contains roughly 100,000 multiple choice questions and 27,000 passages from standardized tests for Chinese students learning English as a foreign language.
    The dataset is split up into: RACE-M, which has approximately 30,000 questions targeted at middle school students aged 12-15, and RACE-H, which has approximately 70,000 questions targeted at high school students aged 15 to 18.
    \citet{Lai2017RACELR} claim that current state of the art neural models at the time of their publishing were performing at 44\% accuracy while the ceiling human performance was 95\%.
    
    \item[AI2 Reasoning Challenge (ARC)] \citep{Clark2018ThinkYH} by Allen Institute for Artificial Intelligence consists of 7,787 grade-school  multiple choice science questions---typically with 4 possible answers.
    The answers generally require external knowledge or complex reasoning.
    In addition, ARC provides a corpus of 14M science-related sentences with knowledge relevant to the challenge.
    However, the training of the models does not have to include, nor be limited to, this corpus.
    
    \item[ReCoRD] \citep{Zhang2018record} contains 12,000 Cloze-style question-passage pairs extracted from CNN/Daily Mail news articles.
    For each pair in this dataset, the question and the passage are selected from the same news article such that they have minimal text overlap---making them unlikely to be paraphrases of each other---but refer to at least one common named entity.
    The focus of this dataset is on evaluating MRC models on their common-sense reasoning capabilities.
    
\end{description}
\section{The MS Marco dataset}
\label{sec:data}

\begin{figure}[t]
\centering
\includegraphics[width=\textwidth]{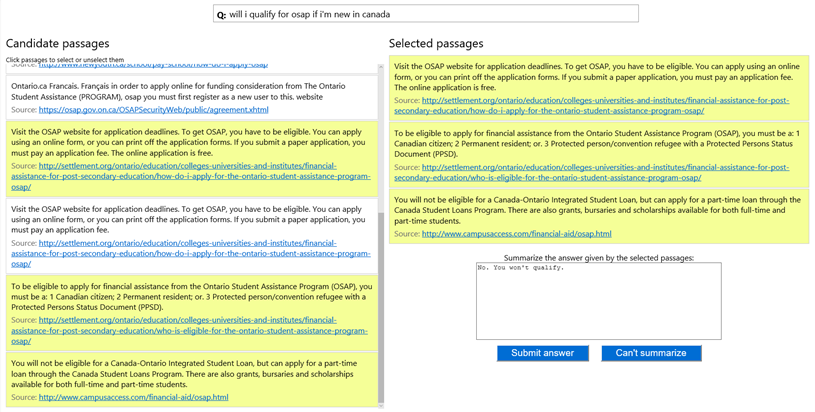} 
\caption{\label{fig:hitapp} Simplified passage selection and answer summarization UI for human editors.}
\end{figure}

To generate the 1,010,916 questions with 1,026,758 unique answers we begin by sampling queries from Bing's search logs.
We filter out any non-question queries from this set.
We retrieve relevant documents for each question using Bing from its large-scale web index.
Then we automatically extract relevant passages from these documents.
Finally, human editors annotate passages that contain useful and necessary information for answering the questions---and compose a well-formed natural language answers summarizing the said information.
Figure \ref{fig:hitapp} shows the user interface for a web-based tool that the editors use for completing these annotation and answer composition tasks.
During the editorial annotation and answer generation process, we continuously audit the data being generated to ensure accuracy and quality of answers---and verify that the guidelines are appropriately followed.

As previously mentioned, the questions in MS MARCO correspond to user submitted queries from Bing's query logs.
The question formulations, therefore, are often complex, ambiguous, and may even contain typographical and other errors.
An example of such a question issued to Bing is: ``in what type of circulation does the oxygenated blood flow between the heart and the cells of the body?''.
We believe that these questions, while sometimes not well-formatted, are more representative of human information seeking behaviour.
Another example of a question from our dataset includes: ``will I qualify for osap if i'm new in Canada''.
As shown in figure \ref{fig:hitapp}, one of the relevant passages include: ``You must be a \begin{enumerate*}
    \item Canadian citizen,
    \item Permanent Resident or
    \item Protected person''.
\end{enumerate*}
When auditing our editorial process, we observe that even the human editors find the task of answering these questions to be sometimes difficult---especially when the question is in a domain the editor is unfamiliar with.
We, therefore, believe that the MS MARCO presents a challenging dataset for benchmarking MRC models.

The MS MARCO dataset that we are publishing consists of six major components:
\begin{enumerate}

    \item \textbf{Questions}: These are a set of anonymized question queries from Bing's search logs, where the user is looking for a specific answer.
    Queries with navigational and other intents are excluded from our dataset.
    This filtering of question queries is performed automatically by a machine learning based classifier trained previously on human annotated data.
    Selected questions are further annotated by editors based on whether they are answerable using the passages provided.

    \item \textbf{Passages}: For each question, on average we include a set of $10$ passages which may contain the answer to the question.
    These passages are extracted from relevant web documents.
    They are selected by a state-of-the-art passage retrieval system at Bing.
    The editors are instructed to annotate the passages they use to compose the final answer as is\_selected.
    For questions, where no answer was present in any of the passages, they should all be annotated by setting is\_selected to $0$.

    \item \textbf{Answers}: For each question, the dataset contains zero, or more answers composed manually by the human editors.
    The editors are instructed to read and understand the questions, inspect the retrieved passages, and then synthesize a natural language answer with the correct information extracted strictly from the passages provided.

    \item \textbf{Well-formed Answers}: For some question-answer pairs, the data also contains one or more answers that are generated by a post-hoc review-and-rewrite process.
    This process involves a separate editor reviewing the provided answer and rewriting it if:
    \begin{enumerate*}[label=(\roman*)]
        \item it does not have proper grammar,
        \item there is a high overlap in the answer and one of the provided passages (indicating that the original editor may have copied the passage directly), or
        \item the answer can not be understood without the question and the passage context.
    \end{enumerate*}
    \eg, given the question ``tablespoon in cup'' and the answer ``16'', the well-formed answer should be ``There are 16 tablespoons in a cup.''.
    
    \item \textbf{Document}: 
    For each of the documents from which the passages were originally extracted from, we include:
    \begin{enumerate*}[label=(\roman*)]
        \item the URL,
        \item the body text, and
        \item the title.
    \end{enumerate*}
    We extracted these documents from Bing's index as a separate post-processing step.
    Roughly 300,000 documents could not be retrieved because they were no longer in the index and for the remaining it is possible---even likely---that the content may have changed since the passages were originally extracted.

    \item \textbf{Question type}: Each question is further automatically annotated using a machine learned classifier with one of the following segment labels:
    \begin{enumerate*}[label=(\roman*)]
        \item NUMERIC,
        \item ENTITY,
        \item LOCATION,
        \item PERSON, or
        \item DESCRIPTION (phrase).
    \end{enumerate*}
    Table \ref{table:query-percentage} lists the relative size of the different question segments and compares it with the proportion of questions that explicitly contain words like ``what'' and ``"where''.
    Note that because the questions in our dataset are based on web search queries, we are may observe a question like ``what is the age of barack obama'' be expressed simply as ``barack obama age'' in our dataset.

\end{enumerate} 

\begin{table}
\caption{Distribution of questions based on answer-type classifier}
\begin{center}
\resizebox{0.5\textwidth}{!}{
\setlength{\tabcolsep}{4pt}
\label{table:query-percentage}
\begin{tabular}{l l}
{Question segment}&{ Percentage of question} \\
\hline
\textbf{Question contains} & \\
YesNo & 7.46\% \\
What & 34.96\% \\
How & 16.8\% \\
Where & 3.46\% \\
When & 2.71\% \\
Why & 1.67\%  \\
Who & 3.33\% \\
Which & 1.79\% \\
Other & 27.83\% \\
\hline
\textbf{Question classification} & \\
Description & 53.12\% \\
Numeric & 26.12\% \\
Entity & 8.81\% \\
Location & 6.17\% \\
Person & 5.78\%  \\
\hline
\end{tabular}}
\end{center}
\end{table}

Table \ref{table:dataset-composition} describes the final dataset format for MS MARCO.
Inspired by \citep{Gebru2018Datasheet} we also release our dataset's datasheet on our website.
Finally, we summarize the key distinguishing features of the MS MARCO dataset as follows:
\begin{enumerate}
    \item The questions are anonymized user queries issued to the Bing.
    \item All questions are annotated with segment information.
    \item The context passages---from which the answers are derived---are extracted from real web documents.
    \item The answers are composed by human editors.
    \item A subset of the questions have multiple answers.
    \item A subset of the questions have no answers.
\end{enumerate}

\begin{table}
\caption{The MS MARCO dataset format.}
\centering
\begin{tabular}{r p{4.1in}} \\ \hline
\textbf{Field} & \textbf{Description} \\ \hline
Query & A question query issued to Bing.\\
Passages & Top 10 passages from Web documents as retrieved by Bing. The passages are presented in ranked order to human editors. The passage that the editor uses to compose the answer is annotated as is\_selected: 1.\\
Document URLs & URLs of the top ranked documents for the question from Bing. The passages are extracted from these documents. \\
Answer(s) & Answers composed by human editors for the question, automatically extracted passages and their corresponding documents. \\
Well Formed Answer(s) & Well-formed answer rewritten by human editors, and the original answer. \\
Segment & QA classification. E.g., {tallest mountain in south america} belongs to the ENTITY segment because the answer is an entity (Aconcagua). \\
\hline
\end{tabular}
\label{table:dataset-composition}
\end{table}

\subsection{The passage ranking dataset}
To facilitate the benchmarking of ML based retrieval models that benefit from supervised training on large datasets, we are releasing a passage collection---constructed by taking the union of all the passages in the MS MARCO dataset---and a set of relevant question and passage identifier pairs.
To identify the relevant passages, we use the is\_selected annotation provided by the editors.
As the editors were not required to annotate every passage that were retrieved for the question, this annotation should be considered as incomplete---\ie, there are likely passages in the collection that contain the answer to a question but have not been annotated as is\_selected: 1.
We use this dataset to propose a re-ranking challenge as described in Section \ref{sec:task}.
Additionally, we are organizing a ``Deep Learning'' track at the 2019 edition of TREC\footnote{\url{https://trec.nist.gov/}} where we use these passage and question collections to setup an ad-hoc retrieval task.

\section{The challenges}
\label{sec:task}
 
 Using the MS MARCO dataset, we propose three machine learning tasks of diverse difficulty levels:
 
 \begin{description}
    \item[The novice task] requires the system to first predict whether a question can be answered based only on the information contained in the provided passages.
    If the question cannot be answered, then the system should return ``No Answer Present'' as response.
    If the question can be answered, then the system should generate the correct answer.
    \item[The intermediate task] is similar to the novice task, except that the generated answer should be well-formed---such that, if the answer is read-aloud then it should make sense even without the context of the question and retrieved passages.
    \item[The passage re-ranking] task is an information retrieval (IR) challenge.
    Given a question and a set of 1000 retrieved passages using BM25 \citep{robertson2009probabilistic}, the system must produce a ranking of the said passages based on how likely they are to contain information relevant to answer the question.
    This task is targeted to provide a large scale dataset for benchmarking emerging neural IR methods \citep{mitra2018introduction}.
\end{description}
\section{The benchmarking results}
\label{sec:experiment}

We continue to develop and refine the MS MARCO dataset iteratively.
Presented at NIPS 2016 the V1.0 dataset was released and recieved with enthusiasm
In January 2017, we publicly released the 1.1 version of the dataset.
In Section \ref{sec:experiment1}, we present our initial benchmarking results based on this dataset.
Subsequently, we release 2.0 the v2.1 version of the MS MARCO dataset in March 2018 and April 2018 respectively.
Section \ref{sec:experiment2} covers the experimental results on the update dataset.
Finally, in October 2018, we released additional data files for the passage ranking task.

\subsection{Experimental results on v1.1 dataset}
\label{sec:experiment1}


We group the questions in MS MARCO by the segment annotation, as described in Section \ref{sec:data}.
The complexity of the answers varies significantly between categories.
For example, the answers to Yes/No questions are binary.
The answers to entity questions can be a single entity name or phrase---\eg, the answer "Rome" for the question what is the capital of Italy".
However, for descriptive questions, a longer textual answer is often necessary---\eg, "What is the agenda for Hollande's state visit to Washington?".
The evaluation strategy that is appropriate for Yes/No answer questions may not be appropriate for benchmarking on questions that require longer answer generation.
Therefore, in our experiments we employ different evaluation metrics for different categories, building on metrics proposed initially by \citep{mitraproposal}. 
We use accuracy and precision-recall measures for numeric answers and apply metrics like ROUGE-L \citep{lin2004rouge} and phrasing-aware evaluation framework \citep{mitraproposal} for long textual answers.
The phrasing-aware evaluation framework aims to deal with the diversity of natural language in evaluating long textual answers.
The evaluation requires several reference answers per question that are each curated by a different human editor, thus providing a natural way to estimate how diversely a group of individuals may phrase the answer to the same question.
A family of pairwise similarity-based metrics can used to incorporate consensus between different reference answers for evaluation.
These metrics are simple modifications to metrics like BLEU \citep{papineni2002bleu} and METEOR \citep{banerjee2005meteor} and are shown to achieve better correlation with human judgments.
Accordingly, as part of our experiments, a subset of MS MARCO where each question has multiple answers is used to evaluate model performance with both BLEU and pa-BLEU as metrics.

\subsubsection{Generative Model Experiments}
The following experiments were run on the V1.1 dataset
\label{sec:gen-model-exp}

Recurrent Neural Networks (RNNs) are capable of predicting future elements from sequence prior.
It is often used as a generative language model for various NLP tasks, such as machine translation \citep{bahdanau2014neural} and question-answering \citep{herman:readCom}.
In this QA experiment setup, we target training and evaluation of such generative models which predict the human-generated answers given questions and/or contextual passages as model input. 

\begin{description}
\item [Sequence-to-Sequence (Seq2Seq) Model.] We train a vanilla Seq2Seq \citep{sutskevar:seq2seq} model with the question-answer pair as source-target sequences.

\item [Memory Networks Model.] We adapt end-to-end memory networks \citep{sukhbaatar2015end}---that has previously demonstrated good performance on other QA tasks---by using summed memory representation as the initial state of the RNN decoder.

\item [Discriminative Model.] For comparison, we also train a discriminative model to rank provided passages as a baseline.
This is a variant of \citep{huang2013learning} where we use LSTM \citep{hochreiter1997long} in place of multi-layer perceptron (MLP).
\end{description}

\begin{table}
\caption{ROUGE-L of Different QA Models Tested against a Subset of MS MARCO}
\begin{center}
\resizebox{\textwidth}{!}{
\setlength{\tabcolsep}{4pt}
\begin{tabular}{>{\raggedright}p{1.25in} >{\raggedright}p{3.2in} c}
\hline
\textbf{} & \textbf{Description} & \textbf{ROUGE-L} \\
\hline
Best Passage & Best ROUGE-L of any passage & 0.351 \\
Passage Ranking & A DSSM-alike passage ranking model & 0.177 \\
Sequence to Sequence & Vanilla seq2seq model predicting answers from questions & 0.089 \\
Memory Network & Seq2seq model with MemNN for passages & 0.119 \\
\hline
\end{tabular}
}
\end{center}
\label{table:model-comparison}
\end{table}

\begin{table}
\caption{BLEU and pa-BLEU on a Multi-Answer Subset of MS MARCO}
\begin{center}
\resizebox{0.4\textwidth}{!}{
\setlength{\tabcolsep}{4pt}
\begin{tabular}{l l l} \hline
\textbf{} & \textbf{BLEU} & \textbf{pa-BLEU} \\ \hline
Best Passage & 0.359 & 0.453 \\
Memory Network & 0.340 & 0.341 \\ \hline
\end{tabular}
}
\end{center}
\label{table:pa-bleu}
\end{table}

Table \ref{table:model-comparison} shows the preformance of these models using ROUGE-L metric.
Additionally, we evaluate memory networks model on an MS MARCO subset where questions have multiple answers.
Table \ref{table:pa-bleu} shows the performance of the model as measured by BLEU and its pairwise variant pa-BLEU \citep{mitraproposal}.

\subsubsection{Cloze-Style Model Experiments} \label{sec:cloze-model-exp}

In Cloze-style tests, a model is required to predict missing words in a text sequence by considering contextual information in textual format.
CNN and Daily Mail dataset \citep{hermann2015teaching} is an example of such a cloze-style QA dataset.
In this section, we present the performance of two MRC models using both CNN test dataset and a MS MARCO subset.
The subset is filtered to numeric answer type category, to which cloze-style test is applicable.

\begin{itemize}
\item \textit{Attention Sum Reader (AS Reader)}: AS Reader \citep{kadlec2016text} is a simple model that uses attention to directly pick the answer from the context.
\item \textit{ReasoNet}: ReasoNet \citep{shen2016reasonet} also relies on attention, but is also a dynamic multi-turn model that attempts to exploit and reason over the relation among questions, contexts, and answers.
\end{itemize}

\begin{table}
\caption{Accuracy of MRC Models on Numeric Segment of MS MARCO}
\begin{center}
\resizebox{0.4\textwidth}{!}{
\setlength{\tabcolsep}{4pt}
\begin{tabular}{c c c} \\ \hline
\textbf{} & \multicolumn{2}{c}{\textbf{Accuracy}} \\
\textbf{} & \textbf{MS MARCO} & \textbf{CNN (test)} \\ \hline
AS Reader & 55.0 & 69.5 \\
ReasoNet & 58.9 & 74.7 \\ \hline
\end{tabular}
}
\end{center}
\label{table:mrc-acc}
\end{table}

\begin{figure}
\centering
\begin{tikzpicture}
	\begin{axis}[
		xlabel=Recall,
		ylabel=Precision,
        grid=major]
	\addplot[color=red,mark=x] coordinates {
        (0.042969,1)
        (0.222656,0.877193)
        (0.289063,0.864865)
        (0.363281,0.827957)
        (0.460938,0.762712)
        (0.554688,0.704225)
        (0.652344,0.664671)
        (0.832031,0.600939)
        (0.941406,0.568465)
        (0.992188,0.555118)
	};
    \addlegendentry{AS Reader}
    
	\addplot[color=blue,mark=o] coordinates {
        (0.074074074,0.999995455)
        (0.222222222,0.924241024)
        (0.296296296,0.874999006)
        (0.383838384,0.842104524)
        (0.437710438,0.830768592)
        (0.535353535,0.773584419)
        (0.653198653,0.716494476)
        (0.801346801,0.663865267)
        (0.909090909,0.625925694)
        (1,0.589225391)
	};
    \addlegendentry{ReasoNet}
	\end{axis}
\end{tikzpicture}
\vspace{-10pt}
\caption{Precision-Recall of Machine Reading Comprehension Models on MS MARCO Subset of Numeric Category}
\label{figure:mrc-pr}
\end{figure}
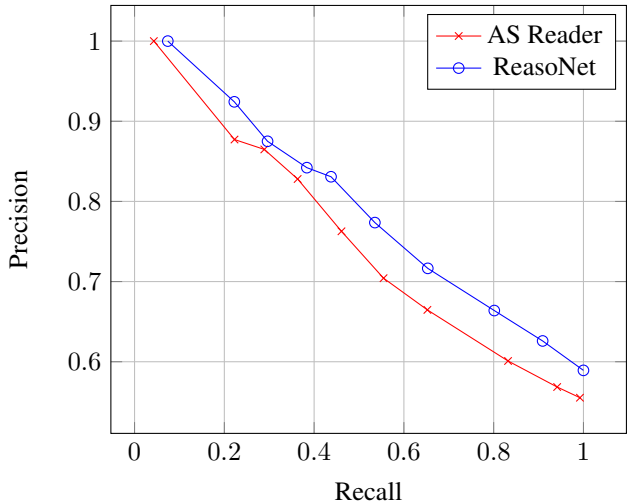

We show model accuracy numbers on both datasets in table \ref{table:mrc-acc}, and precision-recall curves on MS MARCO subset in figure \ref{figure:mrc-pr}.

\subsection{Experimental results on v2.1 dataset}
\label{sec:experiment2}

The human baseline on our v1.1 benchmark was surpassed by competing machine learned models in approximately 15 months.
For the v2.1 dataset, we revisit our approach to generating the human baseline.
We select five top performing editors---based on their performance on a set of auditing questions---to create a human baseline task group. We randomly sample 1,427 questions from our evaluation set and ask each of these editors to produce a new assessment.
Then, we compare all our editorial answers to the ground truth and select the answer with the best ROUGE-L score as the candidate answer.
Table \ref{table:baselines} shows the results.
We evaluate the answer set on both the novice and the intermediate task and we include questions that have no answer.

\begin{table}
\caption{Performance of MRC Span Model and Human Baseline on MS Marco Tasks}
\begin{center}
\resizebox{\textwidth}{!}{
\setlength{\tabcolsep}{4pt}
\begin{minipage}{\textwidth}
\label{table:baselines}
\begin{tabular}{|l|l|l|l|l|l|l|} \hline
{\bf Task}&{\bf ROUGE-L}&{\bf BLEU-1}&{\bf BLEU-2}&{\bf BLEU-3}&{\bf BLEU-4}\\ \hline
BiDaF on Original&0.268&0.129&0.094&0.079&0.070 \\ \hline
Human Ensemble on Novice&0.73703& 0.51586&0.46771&0.43391&0.40540674\\ \hline
Human Ensemble on Intermediate& 0.63044&0.52747&0.45439&0.41235&0.38173\\ \hline
BiDaF on V2 Novice&0.150&0.126&0.094&0.079&0.072\\ \hline
BiDaF on V2 Intermediate&0.170&0.093&0.070&0.059&0.053\\ \hline
\end{tabular}
\end{minipage}
}
\end{center}
\end{table}

To provide a competitive experimental baseline for our dataset, we trained the model introduced in \citep{Clark2017}.
This model uses recent ideas in reading comprehension research, like self-attention \citep{Cheng2016} and bi-directional attention \citep{Seo2016BidirectionalAF}.
Our goal is to train this model such that, given a question and a passage that contains an answer to the question, the model identifies the answer (or span) in the passage.
This is similar to the task in SQuAD \citep{raj:squad}.
First, we select the question-passage pairs where the passage contains an answer to the question and the answer is a contiguous set of words from the passage.
Then, we train the model to predict a span for each question-passage pair and output a confidence score.
To evaluate the model, for each question we chose our model generated answer that has the highest confidence score among all passages available for that question.
To compare model performance across datasets we run this exact setup (training and evaluation) on the original dataset and the new V2 Tasks.
Table \ref{table:baselines} shows the results.
The results indicate that the new v2.1 dataset is more difficult than the previous v1.1 version.
On the novice task BiDaF cannot determine when the question is not answerable and thus performs substantially worse compared to on the v1.1 dataset.
On the intermediate task, BiDaF performance once again drops because the model only uses vocabulary present in the passage whereas the well-formed answers may include words from the general vocabulary.


\section{Future Work and Conclusions}
\label{sec:conclusion}

The process of developing the MS MARCO dataset and making it publicly available has been a tremendous learning experience.
Between the first version of the dataset and the most recent edition, we have significantly modified how we collect and annotate the data, the definition of our tasks, and even broadened our scope to cater to the neural IR community.
The future of this dataset will depend largely on how the broader academic community makes use of this dataset.
For example, we believe that the size and the underlying use of Bing's search queries and web documents in the construction of the dataset makes it particularly attractive for benchmarking new machine learning models for MRC and neural IR.
But in addition to improving these ML models, the dataset may also prove to be useful for exploring new metrics---\eg, ROUGE-2 \citep{Ganesan2018ROUGE2U} and ROUGE-AR\citep{Maples2017TheRA}---and robust evaluation strategies.
Similarly, combining MS MARCO with other existing MRC datasets may also be interesting in the context of multi-task and cross domain learning.
We want to engage with the community to get their feedback and guidance on how we can make it easier to enable such new explorations using the MS MARCO data.
If there is enough interest, we may also consider generating similar datasets in other languages in the future---or augment the existing dataset with other information from the web.

\bibliographystyle{abbrvnat}{
\small
\bibliography{sigproc}  
}
\end{document}